\documentclass[conference]{IEEEtran}
\IEEEoverridecommandlockouts
\usepackage{cite}
\usepackage{verbatim}
\usepackage{amsmath,amssymb,amsfonts}
\usepackage{algorithmic}
\usepackage{graphicx}
\usepackage{textcomp}
\usepackage{xcolor}
\usepackage{array}
\usepackage{float}
\usepackage{siunitx}
\usepackage[format=hang]{caption}
\usepackage[list=true]{subcaption}
\usepackage{hyperref}

\def\BibTeX{{\rm B\kern-.05em{\sc i\kern-.025em b}\kern-.08em
    T\kern-.1667em\lower.7ex\hbox{E}\kern-.125emX}}
\begin{document}

\title{Deep Deterministic Path Following}

\author{
\IEEEauthorblockN{Georg Hess}
\IEEEauthorblockA{
\textit{Chalmers University of Technology}\\
Gothenburg, Sweden \\
georghe@student.chalmers.se}
\and
\IEEEauthorblockN{William Ljungbergh}
\IEEEauthorblockA{
\textit{Chalmers University of Technology}\\
Gothenburg, Sweden \\
willju@student.chalmers.se}
}

\maketitle

\begin{abstract}
This paper deploys the Deep Deterministic Policy Gradient (DDPG) algorithm for longitudinal and lateral control of a simulated car to solve a path following task. The DDPG agent was implemented using PyTorch and trained and evaluated on a custom kinematic bicycle environment created in Python. The performance was evaluated by measuring cross-track error and velocity error, relative to a reference path. Results show how the agent can learn a policy allowing for small cross-track error, as well as adapting the acceleration to minimize the velocity error.
\end{abstract}


\section{Introduction}
Autonomous vehicles have evolved quickly in the last years, mainly thanks to recent development in the field of deep learning. Creating a self-driving vehicle can generally be divided into three subtasks, perception and localization, planning, and control and actuation. This paper is concerned with the latter, namely following a given path, ensuring correct control and actuation of the vehicle. More specifically, a Reinforcement Learning (RL) approach was used for controlling the steering angle $\delta$ and acceleration $a$ of a simulated land vehicle, trying to keep the cross-track error (CTE) and velocity error small. While traditional controlling algorithms exist for this coupled control task \cite{chebly2017coupled} and they provide promising performance, their main drawback is that precise modeling of the vehicle at hand is required. Many parameters must be measured or estimated accurately for the control algorithm to perform adequately. This motivates the use of deep learning where suitable parameters instead are learned by the algorithm itself. Although traditional supervised deep learning can be applied to solve this problem \cite{devineau2018coupled}, this will not be considered here. Instead, Deep Deterministic Policy Gradient will be applied, which is an off-policy learning algorithm suitable to use in a continuous domain where continuous control actions are desirable \cite{lillicrap2015continuous}. The algorithm and its application to the control problem will be explained closer in the coming section.


\section{Related work} \label{sec:related_work}
Reinforcement learning has been successfully applied to various scenarios, such as playing Atari video games \cite{mnih2013playing} and basic control tasks \cite{lillicrap2015continuous}. The basic idea of reinforcement learning consists of having an agent interacting with an environment by each time-step $t$ receiving an observation $x_t$, based on this taking an action $a_t$ and receiving a reward $r_t$. The agent then tries to select actions to maximize the cumulative reward. One of the more commonly known algorithms for finding action policies is called Q-learning, which is a model-free off-policy RL algorithm, introduced already in 1989 \cite{watkins1989learning}. Model-free refers to that no model of the environment is needed to learn suitable actions, while off-policy means that it can learn from previously collected observations. The Q-learning algorithm is value-based, i.e. essentially it uses observations to learn the optimal action for any giving state by keeping track of all possible state-action pairs and continuously evaluating and updating their quality $Q^*(s_t,a_t)$ (hence the name Q-learning). After learning $Q^*(s_t,a_t)$, actions are selected such that $Q^*(s_t,a_t)$ is maximized for each state. However, as the state-space grows larger, the discrete approach used in Q-learning becomes infeasible due to the curse of dimensionality. Furthermore, for a continuous state-space, each state has to be discretized, where there is a trade-off between small discretization steps and keeping low dimensionality. 

The problem of continuous observation spaces can be solved using Deep Q-learning (DQN)\cite{mnih2015humanlevel}. Instead of using a table to store the Q-values, a function approximation based on a neural network is used to estimate the values. In the algorithm a continuous state-vector is fed into a neural network where the output is an estimation of the current Q-value. However, this still leaves one with having to maximize the Q-value over all possible actions, i.e. for each state, one has to search through the actions, finding the one with the highest Q-value. For a large action space or large function approximation network, this becomes computationally costly and time-consuming. 

For handling continuous action spaces, one can use Deep Deterministic Policy Gradient (DDPG) \cite{lillicrap2015continuous} which is an extension of the actor-critic approach in Deterministic Policy Gradient \cite{pmlr-v32-silver14} in combination with findings done by Deep Q-learning \cite{mnih2015humanlevel}. The algorithm is based on two different neural networks, called the actor $\mu(s|\theta^\mu)$ and critic $Q(s,a|\theta^Q)$ networks, where $\theta^\mu$ and $\theta^Q$ denote their weights respectively. The actor network maps vectors $s$ in the state-space into a vector $a$ in the action space, i.e. it uses the knowledge it has about its current state and tries to determine the best possible action to take. The critic network uses both the current state $s$ and the action $a$ taken by the actor-network to decide how good that particular action is to take in this particular state, i.e. trying to approximate the Q-value. The parameters of the critic are updated by using gradient stochastic descent for minimizing the Mean-Squared Bellman Error, similar as in traditional Q-learning. For the actor however, \cite{pmlr-v32-silver14} showed that there exist a policy gradient, i.e. gradient of the policy's performance, along which one updates the parameters of the network
\begin{equation}
    \nabla_{\theta^\mu}J \approx \mathrm{E}_{s\sim\rho^\beta}[\nabla_a Q(s,a|\theta^Q)|_{a=\mu(s|\theta^\mu)}\nabla_{\theta^\mu}\mu(s|\theta^\mu)]
\end{equation}

Deep Q-learning introduced two major ideas to scale Q-learning which are also used in DDPG, namely the use of replay buffers and separate target networks. Replay buffers save a large number of experiences $(s,a,r,s',d)$, where $s'$ denotes the next state and $d$ whether that state is terminal, in a dataset $\mathcal{D}$. One later randomly samples  mini-batches from this dataset, rather than taking the observations directly from the simulation when training the actor and critic networks. By employing a replay buffer, one can suppress the issue that the networks overfit to the most recent data and achieve a more stable training process. Furthermore, it also handles the issues that the observations taken from the simulation are sequentially dependent on each other.

When training the critic network, the Mean-Squared Bellman Error (MSBE) is to be minimized with respect to the parameters $\theta$, i.e. we are trying to minimize the loss function 
\begin{equation}
    \label{eq:bellman-loss}
L = \Big(Q_{\theta}(s, a)-\underbrace{\left(r+\gamma(1-d) \max _{a^{\prime}} Q_{\theta}\left(s^{\prime}, a^{\prime}\right)\right)}_{T}\Big)^2
\end{equation}
by making $Q_{\theta}$ more similar to the target $T$. Here $\gamma$ denotes the discount factor $\gamma \in [0,1]$ determining how much future rewards are valued and $a'$ is the action taken in the next state $s'$. As can be seen in \eqref{eq:bellman-loss}, the target is dependent on the parameters we are trying to learn, namely $\theta$. This can make the training very unstable, leading to the agent not learning anything. Here is where the authors of DQN proposed target networks, which are yet another set of parameters, $\theta_{targ}^Q$ and $\theta^\mu_{targ}$, which lag behind the actual parameters $\theta^Q$ and $\theta^\mu$. In the original DQN paper\cite{mnih2015humanlevel}, the target networks were updated after a fixed number of episodes, however in the DDPG algorithm\cite{lillicrap2015continuous} the authors used Polyak averaging after each training step. The update is therefore described as 
\begin{equation}
    \theta_{targ} \longleftarrow (1-\rho)\theta_{targ} + \rho \theta, \ \rho \ll 1
\end{equation}
By using te target networks instead of the actual networks to calculate the target in \eqref{eq:bellman-loss} one can achieve much more stable training. 

Finally, in RL one has to balance the trade-off between exploration and exploitation. By adding a certain level of randomness in the agent one allows it to explore and learn more about the state-action space before optimizing its policy. In the original DDPG paper \cite{lillicrap2015continuous}, the authors proposed an exploration policy where Gaussian noise was added to the actor-networks output, i.e. directly to the selected action before executing it in the environment. However, as described in \cite{kamran2019learning}, using such independent exploration noise can be very inefficient as a vehicle system acts as a low pass filter for the high-frequency changes in acceleration and steering angle that this type of noise provides. Instead, they proposed an exploration policy where the noise added, $\epsilon_\delta$ and $\epsilon_T$, is sinusoidal with randomly sampled mean and deviation according to
\begin{align}
\epsilon_{\delta}, \epsilon_{a} \sim N\left(\mu, \sigma^{2}\right) \\
\mu_{\delta}=A \sin \left(\omega t+\varphi_{\delta}\right) \\
\mu_{a}=A \sin \left(\omega t+\varphi_{a}\right)
\end{align}
where $\sigma^2$, $A$ and $\omega$ are sampled from a zero-mean Gaussian distribution, while $\varphi_\delta, \varphi_a$ are uniformly sampled in the range $(-\pi, \pi)$. Note here that the values  $\sigma^2$, $A$, $\varphi_\delta, \varphi_a$ and $\omega$ are kept constant over one full episode. 

Furthermore, in \cite{kamran2019learning}, the authors showed that it is advantageous to have an initial explore time (e.g. the first 500 episodes) where the agent, rather than using the output of the actor-network, samples actions uniformly in the range $(-1,1)$. They showed that this drastically improved the performance of the agent and enabled it to learn much better behaviour as it allows for vast exploration in the initial training period.

\section{Method}
Before training the RL agent, an environment must be created with which it can interact. The environment in this project was produced by the authors and consisted of a kinematic bicycle model which took requested steering angles and accelerations as input, and calculated a global position $x$, $y$, velocity $v$, steering angle $\delta$, and heading $\theta$ of the agent. The choice of the kinematic bicycle model is based on the model's simplicity while capturing the overall movements of a car. However, the model was adjusted such that the requested steering angle $\delta_\text{req}$ was limited by a maximum steering rate $\Delta\delta_\text{max}=40$, i.e. the change in steering angle could not exceed a certain threshold. Further, the requested accelerations $a_\text{req}$ were also clipped to lie within adequate limits $[-a_\text{max},a_\text{max}]$ with $a_\text{max} = 5m/s^2$. These changes were introduced to simulate how actuators of a car might behave in real-world conditions. The model can be summarized as selecting a time step $\Delta t$ and vehicle length $L$, then iterating
\begin{align}
    \Delta \delta &= \text{clip}\left(\frac{\delta_\text{req}-\delta_t}{\Delta t},-\Delta\delta_\text{max},\Delta\delta_\text{max}\right) 
    \\
    a_t &= \text{clip}\left(a_\text{req},-a_\text{max}, a_\text{max}\right)
    \\
    x_{t+1} & = v_t \cos(\theta_t) \Delta t\\
    y_{t+1} & = v_t \sin(\theta_t) \Delta t\\
    v_{t+1} & = a_t \Delta t \\
    \delta_{t+1} & = \Delta \delta \Delta t\\
    \theta_{t+1} & = \frac{v_t \tan(\delta_t)}{L} \Delta t
\end{align}

\begin{figure}[tpb]
    \centering
    \includegraphics[width=0.6\linewidth]{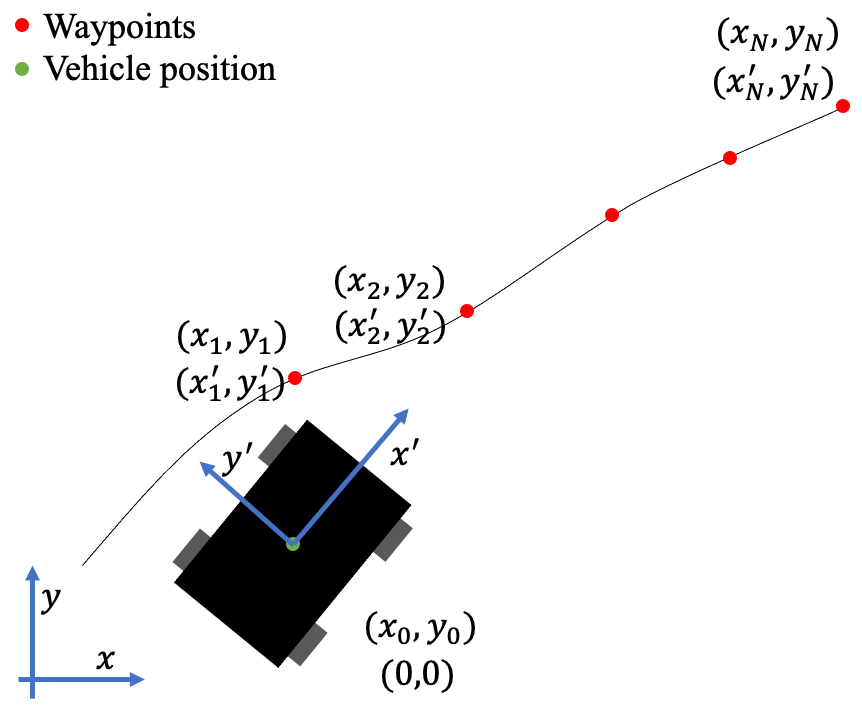}
    \caption{Illustration of state observations for agent.}
    \label{fig:states}
\end{figure}

The model was also used to generate reference paths for the agent to follow. For this, an average velocity $v_{avg} \sim \mathcal{U}(3,20)$ was sampled once before each path generation started and then $\delta_\text{req}$ and $a_\text{req}$ where sampled randomly every time step until the path had a length of $400$m. $\delta_\text{req}$ was sampled from an uniform distribution $\mathcal{U}(-30,30)$ and $a_\text{req}$ was sampled from two different uniform distributions, $\mathcal{U}(-2,2)$ if $v_t > v_{avg}$, otherwise from $\mathcal{U}(0,2)$. The reference trajectory was then processed to consist of waypoints with reference speed every meter. Due to the randomness of the generation, the agent received a unique path to follow each episode. The agent's state observations $s$ consisted of the distance to the 25 nearest future waypoints, expressed in the agent's coordinate frame, along with the difference between current velocity $v_t$ and reference velocity $v_{ref}$ for all 25 waypoints. Additionally, the observed state also included the agent's current velocity and steering angle $v_t$, $\delta_t$. An illustration is shown in Figure \ref{fig:states}, where $(x_i,y_i)$ denote global coordinates and $(x'_i,y'_i)$ coordinates in the agents frame. 

As the agent intends to control the steering angle $\delta$ and the acceleration $a$, the action space is continuous and the actions have been chosen to be in the range of $\delta, a \in [-1, 1]$. Each command must therefore be multiplied with the maximally allowed acceleration and steering angle, as defined by the model described above before it is passed to the simulation.

The reward signal was designed with inspiration from \cite{kamran2019learning} and is described by
\begin{equation}
    R =
    \begin{cases}
      -1 & \text{if $\text{cte} > 0.2$}\\
      1.5 - P_{cte} - P_\delta - P_{v} - P_{acc} & \text{otherwise}
    \end{cases}       
\end{equation}
where, cte is the cross-track error between the agent and the reference path, i.e. perpendicular distance from reference path to agent location, and $P_{cte}$, $P_\delta$, $P_{v}$ and $P_{acc}$ are penalties based on the cross-track error, steering angle, deviation from reference velocity, and the acceleration performed by the agent. The penalties were introduced to minimize cross-track errors, steering angles, velocity errors and jerky maneuvers respectively, and are defined by the following equations
\begin{align*}
    P_{cte} &= 0.8 \text{cte}, \
    P_\delta = 0.1 \delta_t/\delta_{\text{max}}, \\
    P_v &= 0.8 |v_t-v_{ref}| / v_{ref}, \
    P_{acc} = 0.2 |a_\text{req}|
\end{align*}
Additional negative reward was assigned when the normalized velocity error became to large, i.e.
\begin{equation}
    R := R - 1 \ \text{if $|v_t-v_{ref}|/v_{ref} > 0.25$}
\end{equation}
Each simulation was terminated when either the cross-track error $\geq2$ meters, the velocity $v_t \leq 0$ m/s or the agent reached the end of the path. However, only the to first two cases were marked as terminal states in the target $T$.

\begin{figure}[tpb]
    \centering
    \includegraphics[width=0.8\linewidth]{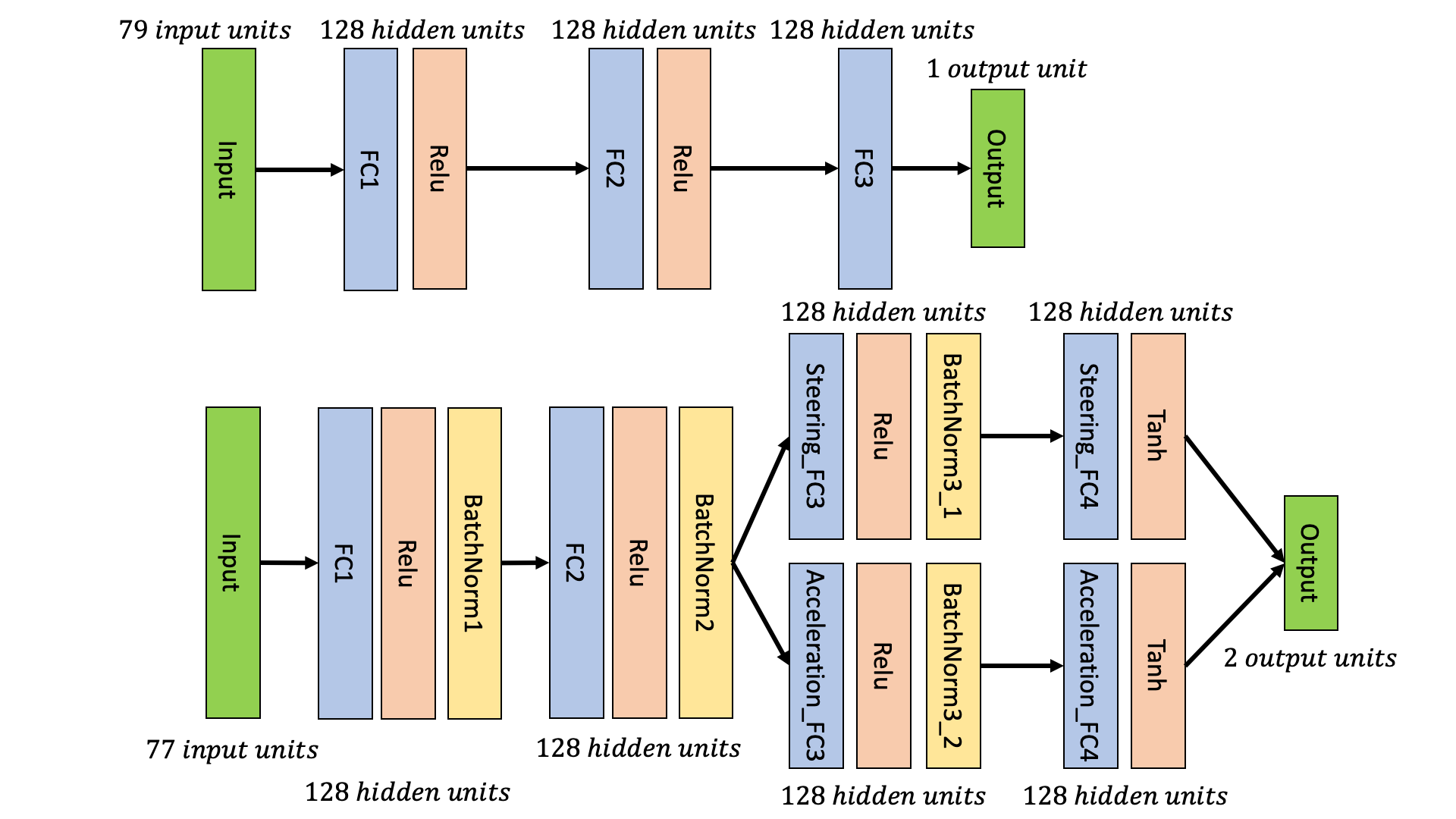}
    \caption{Critic (top) and actor (bottom) network architectures.}
    \label{fig:actor-network}
\end{figure}

As mentioned previously the DDPG architecture consists of two different networks, the actor and the critic. The network architecture for each of the networks are shown in Figure \ref{fig:actor-network}. The actor-network begins with two shared fully connected layer and then branches out into two branches, each producing their own action, i.e. either the steering angle $\delta$ or the acceleration $a$. The two outputs are then concatenated before given as the final output of the actor-network. The weights and biases of the last fully connected layers of network were initialized to $\mathcal{U}(-10^{-6}, 10^{-6})$ as this was found to be crucial for the agent to learn anything at all. 


Furthermore, the replay-buffer size was set to $100,000$ and the sinusoidal exploration policy explained in Section \ref{sec:related_work} was used. To transition smoothly from exploration to exploitation, a noise amplitude multiplier was initialize to 1 and reduced by $0.9996$ every episode. The agent was trained in total for $10$ epochs, each lasting for $500$ episodes. The division into epochs was done mainly for learning rate scheduling.

\section{Results}
\begin{figure}[htpb]
    \centering
    \includegraphics[width=0.7\linewidth]{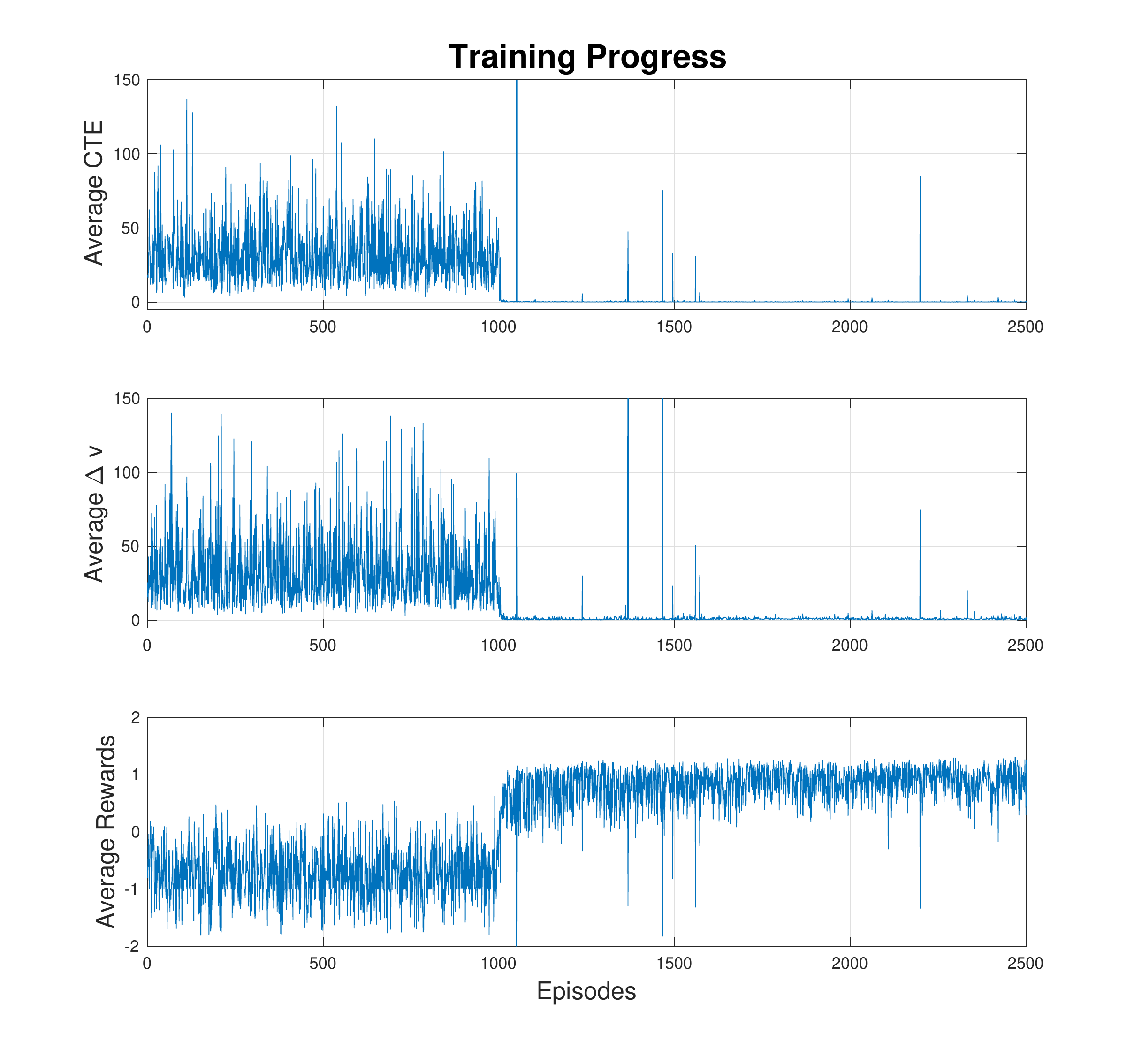}
    \caption{Average cross-track error, reward and velocity error, all normalized with the travelled path length, shown as a function of first 2500 episodes.}
    \label{fig:training-performace}
\end{figure}

In Figure \ref{fig:training-performace} several different metrics from the training are shown for the first 2500 episodes. Starting from the top we see the average cross-track error and velocity error normalized with the travelled path length, as well as the average reward per time-step. After $\sim1000$ episodes a major bump in performance was obtained, which is caused by switching from the aggressive exploration policy used in the initial training phase to the more lenient sinusoidal noise explained in Section \ref{sec:related_work}. Simply from sampling random actions in the action space, the agent was able to learn enough about the environment to perform adequately. When trained without the initial training phase, the agent was not able to perform any satisfactory behaviour after the same number of training steps, thus validating the results in \cite{kamran2019learning}. After the initial bump in performance, the behavior of the agent only changed marginally. While it improved somewhat in all aspects shown in Figure \ref{fig:training-performace}, the biggest change was in consistency. Towards the end of the training the agent was able to maintain a lower variance in all of the aspects discussed above, including the average percent of completion w.r.t. the reference path. 

After the training, the agent was evaluated over a set of 10 test tracks randomly generated using the environment described earlier, where the performance over these is shown in Table \ref{tab:Performace average}.
\begin{table}[tbp]
    \centering
    \caption{Performance over a set of 10 test-tracks}
    \scalebox{0.8}{
    \begin{tabular}{c|c|c|c|c|c}
        Track & Avg cte & Max cte & Avg $\Delta v$ & Max $\Delta v$ & \% of path    \\ \hline
1 & 0.103 & 0.264 & 0.473 & 1.466 & 100.0 \\
2 & 0.141 & 0.32 & 0.465 & 1.813 & 100.0 \\
3 & 0.101 & 0.299 & 0.608 & 2.523 & 100.0 \\
4 & 0.119 & 0.335 & 0.658 & 2.519 & 100.0 \\
5 & 0.113 & 0.641 & 0.62 & 2.34 & 100.0 \\
\textbf{6} & \textbf{0.114} & \textbf{0.464} &\textbf{ 0.752} & \textbf{2.623} & \textbf{100.0} \\
7 & 0.141 & 0.54 & 0.65 & 2.28 & 100.0 \\
8 & 0.107 & 0.31 & 0.609 & 1.937 & 100.0 \\
9 & 0.111 & 0.385 & 0.496 & 1.84 & 100.0 \\
10 & 0.105 & 0.336 & 0.626 & 2.076 & 100.0 \\
\hline
Avg & 0.115 & 0.39 & 0.596 & 2.142 & 100.0
    \end{tabular}}
    \label{tab:Performace average}
\end{table}
From the evaluation one can conclude that the agent is performing passably. The average cross-track error is kept low ($\sim 0.12$ m) while also ensuring that the velocity is kept close to its reference (on average $\sim 0.60$ m/s deviation). For the worst case scenario, the table shows the maximal cross-track error during the 10 evaluations and it never exceeded $0.65$ m. This is important to guarantee that the path following algorithm never strays to far from its reference.  

\begin{figure}[tpb]
    \centering
    \includegraphics[width=0.7\linewidth]{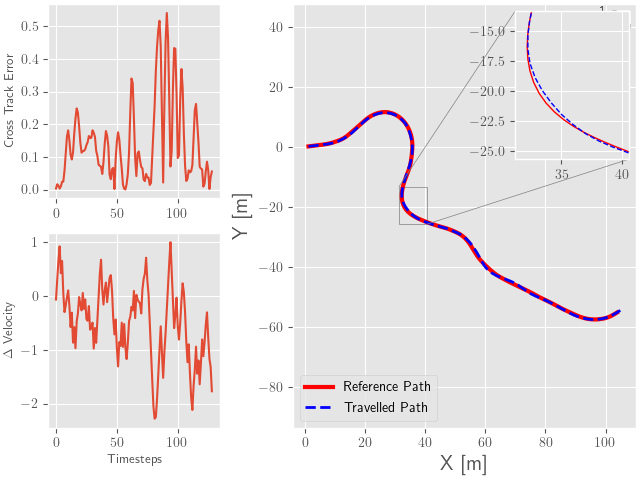}
    \caption{Test Track \#6 with the cross-track error [m] and velocity [m/s] deviation shown over time.}
    \label{fig:example-performace}
\end{figure}

Furthermore, in Figure \ref{fig:example-performace} the evaluation on test track \# 6 (as it is deemed to be representative over the 10 evaluations) is shown in more detail. Here the reference and the actual path travelled is shown together with cross-track error and the velocity error throughout the simulation. As can be seen, the cross-track error rarely exceeds 0.3 m 

\section{Conclusions \& Future Work}
In this paper, we leveraged DDPG to solve the lateral and longitudinal control tasks. While conventionally solved with algorithms such as Pure Pursuit or Model Predictive Control our RL algorithm showed very promising results. The agent used relative distance and velocity to the 25 closest way-points on our reference path to determine the appropriate action. During testing, the agent was able to steer with an average cross-track error of only 0.12m while keeping the agents velocity within 0.6m/s deviation from its reference on average. 

However, it should be noted that the agent was never applied in a real-world setting, but merely a rather simplistic simulator. In future work, it would be interesting not only to apply and test this on an actual vehicle but also to compare it to the performance of the more conventional controllers. 

Furthermore, one could make the simulation more realistic by including e.g. noise in the state observations fed to the agent, or simply use a more sophisticated simulator.

Finally, the authors recognize that the black-box nature of our RL approach raises the issue of how we can guarantee safe behavior at all times. While conventional controllers performance can be analyzed and determined under which circumstances it will perform as intended, our solution is inherently hard to analyze begging the question; can we ever guarantee satisfactory behavior at all times, and if not, under which circumstances can we assure that the RL approach will act appropriately?

\bibliographystyle{IEEEtran}
\bibliography{IEEEabrv,main.bib}

\begin{thebibliography}{1}
\providecommand{\url}[1]{#1}
\csname url@samestyle\endcsname
\providecommand{\newblock}{\relax}
\providecommand{\bibinfo}[2]{#2}
\providecommand{\BIBentrySTDinterwordspacing}{\spaceskip=0pt\relax}
\providecommand{\BIBentryALTinterwordstretchfactor}{4}
\providecommand{\BIBentryALTinterwordspacing}{\spaceskip=\fontdimen2\font plus
\BIBentryALTinterwordstretchfactor\fontdimen3\font minus
  \fontdimen4\font\relax}
\providecommand{\BIBforeignlanguage}[2]{{%
\expandafter\ifx\csname l@#1\endcsname\relax
\typeout{** WARNING: IEEEtran.bst: No hyphenation pattern has been}%
\typeout{** loaded for the language `#1'. Using the pattern for}%
\typeout{** the default language instead.}%
\else
\language=\csname l@#1\endcsname
\fi
#2}}
\providecommand{\BIBdecl}{\relax}
\BIBdecl

\bibitem{chebly2017coupled}
A.~Chebly, R.~Talj, and A.~Charara, ``Coupled longitudinal and lateral control
  for an autonomous vehicle dynamics modeled using a robotics formalism,''
  \emph{IFAC-PapersOnLine}, vol.~50, no.~1, pp. 12\,526--12\,532, 2017.

\bibitem{devineau2018coupled}
G.~Devineau, P.~Polack, F.~Altch{\'e}, and F.~Moutarde, ``Coupled longitudinal
  and lateral control of a vehicle using deep learning,'' in \emph{2018 21st
  International Conference on Intelligent Transportation Systems (ITSC)}.\hskip
  1em plus 0.5em minus 0.4em\relax IEEE, 2018, pp. 642--649.

\bibitem{lillicrap2015continuous}
T.~P. Lillicrap, J.~J. Hunt, A.~Pritzel, N.~Heess, T.~Erez, Y.~Tassa,
  D.~Silver, and D.~Wierstra, ``Continuous control with deep reinforcement
  learning,'' \emph{arXiv preprint arXiv:1509.02971}, 2015.

\bibitem{mnih2013playing}
V.~Mnih, K.~Kavukcuoglu, D.~Silver, A.~Graves, I.~Antonoglou, D.~Wierstra, and
  M.~Riedmiller, ``Playing atari with deep reinforcement learning,''
  \emph{arXiv preprint arXiv:1312.5602}, 2013.

\bibitem{watkins1989learning}
\BIBentryALTinterwordspacing
C.~J. C.~H. Watkins, ``Learning from delayed rewards,'' Ph.D. dissertation,
  King's College, Cambridge, may 1989. [Online]. Available:
  \url{http://www.academia.edu/download/50360235/Learning_from_delayed_rewards_20161116-28282-v2pwvq.pdf}
\BIBentrySTDinterwordspacing

\bibitem{mnih2015humanlevel}
\BIBentryALTinterwordspacing
V.~Mnih, K.~Kavukcuoglu, D.~Silver, A.~A. Rusu, J.~Veness, M.~G. Bellemare,
  A.~Graves, M.~Riedmiller, A.~K. Fidjeland, G.~Ostrovski, S.~Petersen,
  C.~Beattie, A.~Sadik, I.~Antonoglou, H.~King, D.~Kumaran, D.~Wierstra,
  S.~Legg, and D.~Hassabis, ``Human-level control through deep reinforcement
  learning,'' \emph{Nature}, vol. 518, no. 7540, pp. 529--533, Feb. 2015.
  [Online]. Available: \url{http://dx.doi.org/10.1038/nature14236}
\BIBentrySTDinterwordspacing

\bibitem{pmlr-v32-silver14}
\BIBentryALTinterwordspacing
D.~Silver, G.~Lever, N.~Heess, T.~Degris, D.~Wierstra, and M.~Riedmiller,
  ``Deterministic policy gradient algorithms,'' ser. Proceedings of Machine
  Learning Research, E.~P. Xing and T.~Jebara, Eds., vol.~32, no.~1.\hskip 1em
  plus 0.5em minus 0.4em\relax Bejing, China: PMLR, 22--24 Jun 2014, pp.
  387--395. [Online]. Available:
  \url{http://proceedings.mlr.press/v32/silver14.html}
\BIBentrySTDinterwordspacing

\bibitem{kamran2019learning}
D.~Kamran, J.~Zhu, and M.~Lauer, ``Learning path tracking for real car-like
  mobile robots from simulation,'' in \emph{2019 European Conference on Mobile
  Robots (ECMR)}.\hskip 1em plus 0.5em minus 0.4em\relax IEEE, 2019, pp. 1--6.

\end{thebibliography}

\end{document}